# SITUATE - Synthetic Object Counting Dataset for VLM training

René Peinl, Vincent Tischler, Patrick Schröder, Christian Groth
*Institute of Information Systems, Hof University of Applied Sciences*
{ rene.peinl, vincent.tischler.3, patrick.schroeder, christian.groth}@hof-university.de

Keywords: Vision language model, counting, spatial reasoning, computer vision.

Abstract: We present SITUATE, a novel dataset designed for training and evaluating Vision Language Models on counting tasks with spatial constraints. The dataset bridges the gap between simple 2D datasets like VLMCountBench and often ambiguous real-life datasets like TallyQA, which lack control over occlusions and spatial composition. Experiments show that our dataset helps to improve generalization for out-of-distribution images, since a finetune of Qwen VL 2.5 7B on SITUATE improves accuracy on the Pixmo$_{count}$ test data, but not vice versa. We cross validate this by comparing the model performance across established other counting benchmarks and against an equally sized fine-tuning set derived from Pixmo$_{count}$.

## 1 INTRODUCTION

Vision-language models (VLMs) have demonstrated remarkable capabilities in connecting visual understanding with natural language reasoning. They perform well on tasks like image captioning, visual question answering, and cross-modal retrieval, where the goal is to relate what is seen in an image to a corresponding text.

However, despite these advances, VLMs continue to struggle with tasks that require precise quantitative reasoning, particularly object counting (Jiang et al., 2023) and spatial relationship. While they can describe a scene coherently and identify relevant objects, their claims regarding quantity of objects or spatial relations are often unreliable or inconsistent. They recognize objects and scenes well but struggle to translate this understanding into precise numbers, especially when object count goes beyond five. It seems similar to humans who can recognize small object numbers at a glance, but need to explicitly count when numbers get higher. The latter is missing in VLMs.

When it comes to counting, VLMs also fail to deliver accurate results due to inherent biases, where they remember elements from training (Zhang & Wang, 2024). Vo et. al. (Vo et al., 2025) found an accuracy of only 17 % if real count deviates from usual count, e.g. a hen has three legs instead of two. When it comes to color counting all VLMs of different size show poor performance (Liang et al., 2025).

These weaknesses limit their use in situations where exact quantities or precise spatial understanding are required, like inventory management.

Our goal is therefore to provide a dataset that can be used to finetune VLMs to improve their counting abilities, especially with respect to additional conditions like colors or spatial position.

Our main contribution is a new dataset that can be used for object counting and complements existing datasets in valuable aspects. Further, our analysis gives deeper insight into how VLMs learn counting.

The remainder of the paper is structured as follows. We first discuss related work and derive some learnings for our own dataset and approach. Afterwards, we describe how we constructed our dataset. Then we introduce the experimental setup for finetuning a VLM and comparing it with other datasets and models before listing the results. We then discuss our findings and finish with conclusions and future work.

## 2 RELATED WORK

Guo et. al. (Guo et al., 2025) present a comparative study which reveals that all tested VLMs show poor performance at counting up to 20 as soon as multiple shapes are involved. The models tested include open weight models like Llama 4 Maverick and GLM 4.5V but also leading proprietary models like GPT-4o and Gemini 2.5 flash. The best model achieved an accuracy of 60%. They introduce a relative error measure that computes the percentage of deviation from the ground truth. The best models' relative error is between 4% and 5%. TallyQA uses RSME instead, the root mean squared error, to provide a second quality measure (Acharya et al., 2019) besides accuracy. RSME punishes larger errors more and is independent of the absolute value of the ground truth. The dataset from (Guo et al., 2025) was not made publicly available.

Chain-of-thought counting is used in models which always expose internal reasoning, like in Meta's Llama (META, 2025) or Kimi-VL (Team et al., 2025). Structured prompting usually incorporates a prompt that first lists all objects and then counts them in a second step (Lin & Chan, 2024) or uses a so called semantic visual prompt tuning (Zhao et al., 2025). Recent VLMs like Qwen3-VL (Alibaba Qwen Team, 2025) are heavily trained with this style which leads to a verbose output increasing token count, but potentially also increasing accuracy for larger numbers and structured scenes with groups of objects.

## 2.1 Training and finetuning

Additional information strategies employ exemplar crops (Xu et al., 2023) or additional detection masks for different regions (Qharabagh et al., 2025) or even adding a fuzzy reward model (Wang et al., 2025), which is specifically designed for much larger numbers without the need for precision.

While these approaches improve counting performance to some extent, they do not enable the model to develop a true understanding of numerosity. In contrast to that, fine-tuning a vision-language model with an adapted loss function (Paiss et al., 2023) or a dedicated counting dataset addresses the problem directly. Paiss et al. (Paiss et al., 2023) demonstrate that with a CLIP model. They show that even for small numbers, the baseline CLIP model has significant problems (see Figure 1).

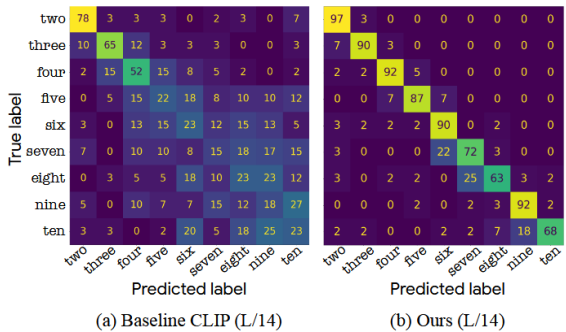

Figure 1: Confusion matrix for different CLIP models (Paiss et al., 2023)

Fine-tuning the model further only improves the accuracy on their CountBench test dataset from 33% to 50% with a mean deviation of one. Only after strict filtering of the dataset derived from LAION-400M and balancing the training data, which they don't publish, for relatively equal number of big and small numbers lead to an accuracy of 76%.

On the other hand the problem of poor spatial understanding has also been addressed by adapting foundation model architectures and finetuning on specialized datasets with extended positional information (Dorkenwald et al., 2024; Huang et al., 2024). On the downside the counting aspect is not represented in these approaches.

## 2.2 Existing counting datasets

The most relevant datasets for this task are Pixmo count (Deitke et al., 2024) and TallyQA/HowMany-QA (Acharya et al., 2019), which are derived from the COCO and HumanGenome datasets. Although these datasets provide some information about counting, we will show in chapter 3, that they have significant problems. Besides this, the spatial information is very limited.

ColorBench includes a small portion (103 of 5885 images) of the benchmark concerned with counting of objects with a specific color (Liang et al., 2025). All tested models including big commercial ones like GPT-4o perform bad with Llava-OV-72B scoring best with 50.5% accuracy compared to a human baseline of 81.3%. The answers are multiple choice with five choices and a mapping from numbers to letters, e.g., A=7, B=5, …, expected answer = C. This seems unnecessary and potentially compromising the construct validity. Furthermore, the real-world images partially include heavy occlusion and somehow ambiguous questions (see Figure 2) that partially explain the low accuracy.

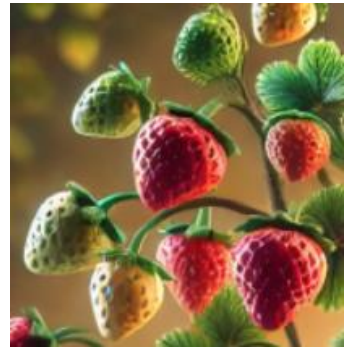

Figure 2: Example image from ColorBench (cropped). question: How many berries are green? 1,2,3 or 4?

Synthetic data is recently gaining interest in LLM and VLM training, e.g. for personalized question answering (Braga et al., 2024) or to enhance logical reasoning skills as shown in the Phi 4 mini model (Microsoft et al., 2025). Wu et al. (Wu et al., 2025) are using it to stabilize continual training of VLMs and avoid catastrophic forgetting. They find that their Stable Diffusion based method outperforms other approaches when applied on CLIP. They achieve this with only 1,000 synthetic images compared to other approaches using 100,000 images from ImageNet. Lee at al. (Lee et al., 2025) create synthetic data to address shortcomings in VLMs for low-resource languages. They test their approach on text recognition tasks for the Korean language and find it effective.

Currently there is no approach that tackles the problems of counting and spatial understanding together in a unified way. Therefore, we propose a new dataset that addresses these problems specifically. We verify the effectiveness by using it to finetune a foundation model and validate the results.

## 3 DATASET

In this chapter, we outline the motivation for a new dataset and then explain its creation.

### 3.1 Motivation

The existing object counting dataset **Pixmo count** by AllenAI uses real-life images and features a single object category per image. It consists of 36,900 images for training and 540 images for eval and test each. 5,000 images in the training set are adversarial questions that ask for objects not present in the image. Another 24,600 images have an object count of 1-5. 7,058 images show six to 15 objects, which is our focus area and 255 images show more than 15 objects, only 79 of them more than 20 objects. The dataset has 365 different object classes, with people accounting for 2/3 of all images. Airplanes, birds, dogs and cats follow with a large gap and an image count of 800-500 each. Most other object classes have less than 10 images.

**CountBench** has published only the test dataset with 540 images evenly distributed across the number classes 2-10. Although the pictures are derived from LAION-400M, they still contain many artificial contents like icons, especially for larger number classes. For the number nine, e.g., there is also a strong bias towards a visual pattern (see Figure 3), so it is highly likely that models trained on such data will not learn counting but identifying the visual pattern. This can also be seen in the confusion matrix that show much higher correct results for nine than eight.

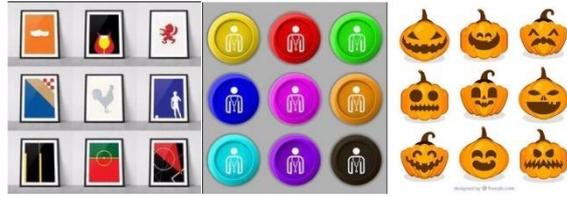

Figure 3: Example images from CountBench for number 9

**TallyQA** is a large-scale counting dataset already created in 2019 with nearly 288,000 real-world image / question pairs (Acharya et al., 2019). It contains not only simple counting questions, but also more complex ones that constrain the objects to count with additional attributes, e.g. counting only those palm trees in the background of a tennis game photo, that are situated between two light posts. It also contains a number of adversarial questions that ask for things not present in the image, e.g. black cows in a picture that shows three brown cows. Their focus is on smaller numbers as visible in Figure 4. However, due to the large scale of the dataset, there is still a considerable amount of images left with ground truth answer 5 and above.

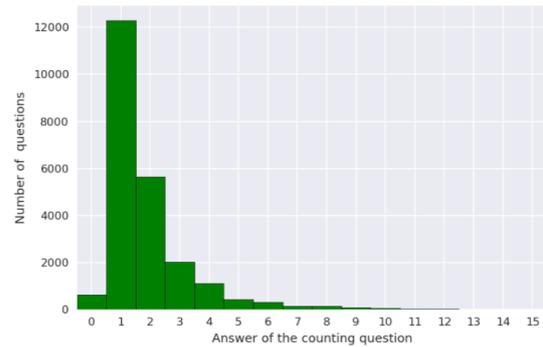

Figure 4: distribution of ground truth answers for TallyQA test simple dataset (Acharya et al., 2019)

In contrast, **our dataset** has only four object classes, cubes, spheres, cones and cylinders and consists of 3D renderings created in Blender. It features multiple object classes (usually two) per image and additionally uses colours of the objects and location in relation to a table to make counting more challenging, similar to the complex questions in TallyQA. It is also similar to VLMCountBench (Guo et al., 2025) but uses high quality 3D renders instead of simple 2D graphics and is publicly available. The scene has varied backgrounds and floor textures, but besides that always uses a similar layout with a table in the center and objects on, under, in front, to the left or right of the table. We also use adversarial questions and pay attention to ask for objects that are close to what is seen in the scene, e.g. for green cubes, if there are blue cubes and green spheres in the image. Our focus is on 5-15 objects per image, but the dataset also contains smaller counts and can be easily expanded for larger numbers, if you decrease the size of a single object in the scene.

Table 1: comparison of object counting datasets

|  | SITUATE | Count Bench | Tally QA | Pixmo |
|---|---|---|---|---|
| Size train / test | 22,807 +496 | 491 | 288k + 498 | 23,342 + 527 |
| Test # | 0-15 | 1-10 | 0-15 | 2-10 |
| Train # | 0-15 | unclear | 0-15 | 0-254 |
| Objects | 4 types | 100+ | 100+ | 5+300 |
| Type | Synth | Photo | Photo | Photo |

## 3.2 Dataset Generation

For the creation of the synthetic dataset, we employ Blender, an open-source 3D software with a good Python API, which gives programmatic access to all aspects of the software. Additionally, we utilize BlenderProc (Denninger et al., 2023), a pipeline built on top of the Blender API specifically designed for the procedural generation and rendering of synthetic datasets.

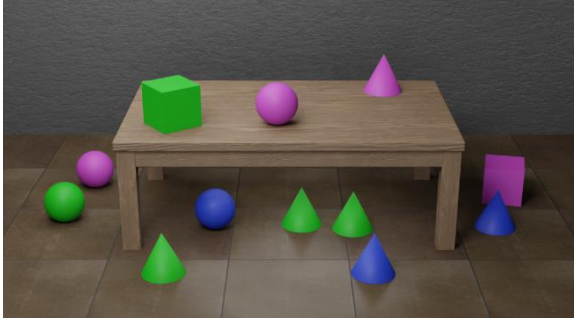

Figure 5: example image for our dataset

All parameters for the image generation are stored in a JSON configuration file, which includes
- Image properties, such as output resolution, the sampling rate used during rendering, and the number of rendered camera perspectives.
- Environmental factors, including ranges for lightning strength, room dimensions, and overall scene layout.
- Object-level attributes, such as the number of objects, their size ranges, and the likelihood of their placement in different locations.

These parameters create a configuration space which we sample to create the dataset. The base scene for the dataset consists of a table placed inside a room.

Table 2: example for the questions in SITUATE

| type | shape.count |
|---|---|
| question | How many cones can be seen in the image? |
| numeric gt | 6 |
| short gt | There are 6 cones in the image. |
| verbose gt (ground truth) | Let´s analyze the scene! On top of the table, I can see 2 cones. On the ground to the right of the table, I can see 3 cones. On the ground in front of the table, I can see 1 cone. In total there are 6 cones in the image! |

**Object placement** within these locations is randomized according to probabilities set in the configuration. To position an object, the bounding box of the table is retrieved. The x and y coordinates are sampled uniformly within the table's horizontal bounds, while the z coordinate depends on the assigned location. To avoid object overlapping and occlusions that might hinder clear visual understanding, the table's horizontal extent along the x-axis is divided into equally sized bins. By placing each object randomly within its own dedicated bin, we ensure that objects are distributed into distinct, non-overlapping regions, effectively reducing the chances of visual confusion. The output images are saved as PNG. Resulting images are shown in Figure 5. To ensure the quality of the rendered images, we implemented a **validation** process addressing common issues such as low contrast between objects and their background. Using segmentation maps, object pixels are isolated and then dilated to include neighboring background pixels. We compute the mean colors in the Lab-colorspace for both the object and its surrounding background, then measure their color difference as $\Delta E$:

$$\Delta E = \sqrt{\left(L_{\text{obj}} - L_{\text{bg}}\right)^2 + \left(a_{\text{obj}} - a_{\text{bg}}\right)^2 + \left(b_{\text{obj}} - b_{\text{bg}}\right)^2}$$

A large $\Delta E$ indicates the object clearly stands out from the background. In our case, a $\Delta E$ value below 12.5 was empirically determined to be an insufficient contrast. In these cases, materials for all primitives and environmental elements are reassigned randomly, and then the scene is re-rendered.

**Visual question answering data**

To create a dataset that can be used for finetuning the counting capabilities in a visual question answering setup, we need to add textual descriptions and questions to the images. To achieve this, comprehensive metadata, including the number of elements per color, per shape, per location and in total were used. Additionally, the exact location of the shapes is stored as well as the full raw scene description, including object properties, spatial arrangements, and materials.

Table 3 statistics about the SITUATE training dataset

| gt answer | items | gt answer | items |
|---|---|---|---|
| 0-5 | 3,100 each | | |
| 6 | 2,277 | 11 | 470 |
| 7 | 1,358 | 12 | 441 |
| 8 | 842 | 13 | 442 |
| 9 | 545 | 14 | 453 |
| 10 | 479 | 15 | 445 |

This whole data is used to form one of 6 question types: color, shape, location, object, composite and adversarial. A question of each kind is created per image with templates for each class to represent a variety of phrasing. Three different types of ground truths are provided for each image-question pair as seen in Table 2. The overall dataset consists of 23,252 image-question-answer triplets, where each image has a resolution of 1024 x 576 pixels. The image feature 5–15 object instances. Each scene is rendered from 5 different camera perspectives, which results in a total of 6.875 images. For each image there are 4 different questions, resulting in 27,500 entries, which were filtered to the final 23,252.

## 4 EXPERIMENTS

To demonstrate the quality of our dataset we finetune a Qwen 2.5 VL 7B model. We chose this model as a basis since it demonstrated its quality with good results in many benchmarks in the past including our own German VLM benchmarks ((Peinl & Tischler, 2025a), (Peinl & Tischler, 2025b)).

As a reference for current state of the art VLMs we chose the recent Qwen 3 VL 32B that was released in October 2025 and offers 99.4% of the accuracy of the biggest Qwen 3 VL 235B A22B across all benchmarks reported by Alibaba. Furthermore, we consider Molmo 7B-D0924. Although it is an older model (release in September 2024), it is the only model known to have been trained on the Pixmo count dataset, so it may have an advantage in our benchmark.

We developed four fine-tuned variants of the QwenVL-2.5-7B-Instruct model (baseline): Verbose, Non-verbose, Pixmo, and Mixed (a combination of our dataset and Pixmo). For the Verbose and Non-verbose variants, we used 23.252 image / question / answer triples from both of our dataset version. The Pixmo-Subset was selected to be as similar as possible to ours regarding the groundtruth numbers. Finally, we fused the two datasets to create a unified one with roughly the same number of items by reducing the images with lower object count and keeping the big ones, so the result is more balanced.

We are using parameter-efficient finetuning with the unsloth framework (Han et al., 2023) and Low-Rank Adaptation (LoRA). In preliminary tests, we determined that rank 16 and an alpha value of 32 as good settings for our case. We train the model in the different experiments for 1 epoch each on the 23,252 items with a batch size of 4 and gradient accumulation steps set to 4 on a single Nvidia Pro 6000 GPU. As optimizer we use adamw_8bit, set the learning rate to 1e-4 and the warmup to 300. The resulting model with adapter is evaluated on the held-out 496 item test split of our dataset, that consists of 31 questions for each answer category of 0-15, as well as the Pixmo count test dataset, CountBench and TallyQA. The latter one is a filtered subset of TallyQA that includes all the questions with answers above from 11-15 and 30 questions each marked as simple or complex for the classes 5 ... 10, summing up to 508 questions altogether. However, we found some questions to be rather unfair, since it is not clear that mostly occluded objects should be counted as well. For example, one question asks for motorcycle, but at least one of the motorcycles that was counted for the ground truth is not visible at all and can only be deducted from the head of a person that looks like a motor cyclist and is presumably riding a motorcycle and not just standing on the street.

All datasets are published on Huggingface. https://huggingface.co/iisys-hof/datasets

We trained the models only for one epoch to reach the same amount of compute for all experiments. Our goal was not to reach the best possible model quality but investigate how VLMs learn to count.

## 5 RESULTS

As seen in Table 4, by finetuning the baseline on the different versions of the dataset, we observe that both the non-verbose and the Pixmo finetune decrease the accuracy on our SITUATE test data. The verbose finetune increases the accuracy to 50% and the mixed finetune adds another 2.5% accuracy. Unsurprisingly, the Pixmo finetune performs best on the Pixmo test data, but the non-verbose finetune also leads to a moderate increase in accuracy compared to the baseline. The verbose finetune decreases it.

Table 4: results of the various finetunes and the baseline

| Test \ train | Verbose | Baseline | Non-verbose | Pixmo | Mixed |
|---|---|---|---|---|---|
| SITUATE | 50.3% | 36.3% | 29.5% | 27.3% | 52.7% |
| Pixmo count | 35.6% | 51.8% | 54.4% | 62.7% | 60.0% |
| CountBench | 47.1% | 81.9% | 67.4% | 85.1% | 81.9% |
| TallyQA | 21.7% | 25.5% | 24.3% | 28.7% | 27.5% |
| Average | 38.5% | 48.9% | 43.9% | 51.0% | 55.5% |

For CountBench, the baseline is already quite good with nearly 82% accuracy. The Pixmo finetune can increase that moderately (+3%). TallyQA is the opposite and extremely hard, even for humans and also contains ambiguities and heavy occlusion. Pixmo contains much easier examples and therefore the accuracy of the baseline increases moderately (+3%). The SITUATE dataset seems to be too different to provide positive training impulse and decreases accuracy. Especially the verbose output does not help here. Experiments with refined prompt formats like "count the objects on the left half first and then the one on the right" were already shown to lead to worse results (Guo et al., 2025).

Compared to the other models (see Table 5), our best finetune (mixed) with just a single epoch of finetuning achieves an increase of 6.6% across all four datasets compared to the baseline. Molmo 7B-D performs exceptionally well on the Pixmo count dataset, it was trained on and beats Qwen3 VL 32B by 13%. It is also very strong on CountBench and even the hard TallyQA. However, for SITUATE it is only moderately better than the Qwen 2.5 VL baseline and beaten by far by Qwen3 VL 32B. The latter is the only model that shows good generalization capabilities, which becomes especially prominent when looking at the median instead of the average (77.1% compared to 67.2% for Molmo)

Table 5: Comparison of our finetune with other VLMs

| Test data | Qwen2.5-VL 7B | Ours (Mixed) | Molmo-7B-D | Qwen3 VL 32B |
|---|---|---|---|---|
| SITUATE | 36.3% | 52.7% | 47.4% | 79.2% |
| Pixmo count | 51.8% | 60.0% | 87.1% | 75.0% |
| CountBench | 81.9% | 81.9% | 92.7% | 92.3% |
| TallyQA | 25.5% | 27.5% | 37.8% | 33.1% |
| Average | 48.9% | 55.5% | 66.2% | 69.9% |

# 6 DISCUSSION

Our dataset fills a gap and is a valuable addition to the existing datasets. CountBench is too easy for VLMs in 2025, presumably because of much similar data in the training datasets. Our baseline scored already over 80%. TallyQA is on the other hand much too hard and ambiguous in many cases, which means that it is not helpful to discriminate between VLMs that are good or bad at counting. The difference between our baseline and Qwen3 32B is merely 8% absolute. The SITUATE dataset also contains test data for bigger numbers up to 15 whereas the three other datasets just provide data up to 10.

Using the $Pixmo_{train}$ dataset to finetune our baseline, leads to an increase of 11% absolute on $Pixmo_{test}$ but to a drop in accuracy for our dataset. Looking at the number classes, we observe a slight increase in accuracy for the numbers four and five, but worse results in all other cases.

The other way around, using our dataset to finetune the baseline leads to a small increase of 2.5% absolute on the $Pixmo_{test}$ data with the non-verbose finetune and a 7.5% absolute increase on our own dataset for the verbose finetune.

Our analysis shows that verbose and non-verbose style have their own strengths and weaknesses. While the verbose style is helpful for results of 5 and above (improvement of the finetune compared to both the baseline as well as the non-verbose finetune), it actually leads to hallucinations for smaller numbers. The finetune e.g. makes up two additional blue cones when asked for blue cubes and answers with "three blue objects". The model tries to find something to add together and if it doesn't find it with the objects mentioned in the question, it hallucinates.

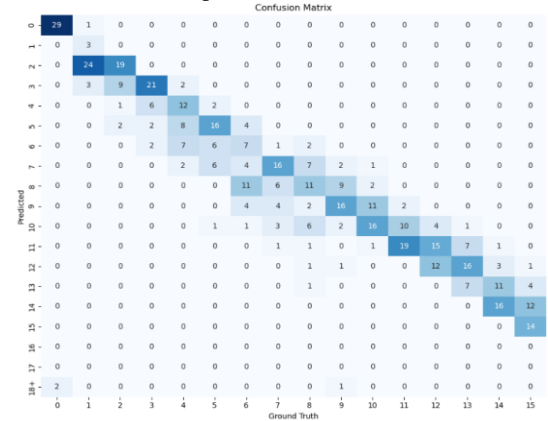

Figure 6: confusion matrix for the verbose finetune

Interestingly, for the Pixmo dataset, where it is hard to find suitable subgroups of items so that the strategy of the verbose training cannot work well, the verbose finetune is still closer to the correct result than the non-verbose, although accuracy is considerably lower. For the numbers five to nine, the verbose finetune maintains a difference around 1.7, whereas the non-verbose model has a deviation of 2.1. Nevertheless, the accuracy of the non-verbose model is 54.6%, compared to not even 36% for the verbose finetune.

Another surprising observation is that the non-verbose finetune makes considerably more mistakes than the verbose finetune on the Pixmo test dataset by stating that it can see none of the objects that is asked for (50x compared to 8 times out of 527). It kind of overfitted on our adversarial questions in the training data and even hallucinated an answer that deals with the color of objects instead of the type that was asked for. Since 100% of the adversarial questions were answered correctly already by the baseline, it seems to be harmful to further train the model on this.

Overall, the mixed dataset consisting of Pixmo and Verbose SITUATE training data achieves the best results across all benchmarks and leads to no degradation of the base model in any of the tests. We speculate that we will be able to further increase it by using the non-verbose training data for the numbers up to four and the verbose style for five and above.

When comparing the confusion matrices shown in Figure 6 and Figure 7, it becomes obvious that the verbose finetune significantly reduces the variance for the numbers nine and above. Wrong counts tend to be lower than they should be. In contrast to that, the baseline results are scattered in both directions, although also with a trend to lower numbers.

We noticed a mysterious weakness of our baseline for the number six. The accuracy drops from 39% for the number five to 19% for the number six, just to go up again to 35% for the number seven. This weakness at the number six is still present in our verbose and non-verbose finetunes but can be cured with the mixed dataset of half Pixmo, half SITUATE training data (61% accuracy). We hypothesize that the base model was trained on number 0-5. Since six is similar enough to five, the model "thinks" it should answer 5 (24 out of 31). Seven looks different enough, so that accuracy goes up again.

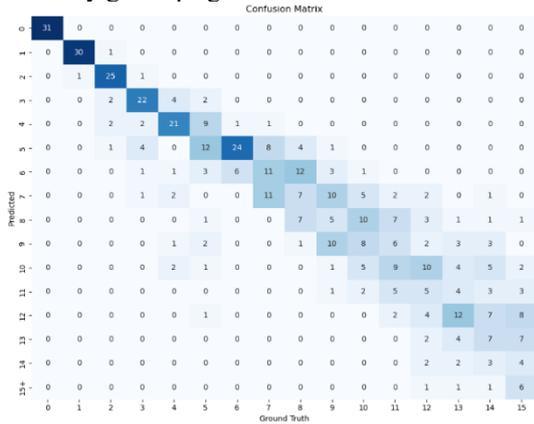

Figure 7: confusion matrix for the baseline on SITUATE

The slight increase in accuracy for the number nine compared to eight is amplified with the verbose finetune to 55% (+20% over eight) and even 77% with the mixed dataset (+39% over eight). In contrast to the visual pattern noticed for CountBench, neither Pixmo count nor SITUATE test data show any noticeable patterns for the number nine.

## 7 LIMITATIONS

Our benchmark currently is specific to a single setup with a table and basic geometric shapes. We showed, that models finetuned on this benchmark generalize to different types of images to a certain degree. However, having a more diverse set of scenes would be beneficial, although we made the scene as diverse as possible by adding multiple camera angles and varying the floor and wall textures.

We only finetuned a single VLM, Qwen 2.5 VL 7B. Although we are relatively sure that other VLMs will benefit equally from being finetuned on our dataset, we have not explicitly tested that.

We finetuned only for a single epoch to avoid overfitting and because reaching the best quality was not the goal of the study. For the future, we still plan on training multiple epochs to analyze results on the resulting models as well.

We did not determine a human baseline for our benchmark, but we are relatively sure, that average humans are able to achieve 90% or even 95% accuracy, since we avoided all the problems discussed for other datasets.

## 8 CONCLUSION AND OUTLOOK

We presented SITUATE, a new dataset for training and evaluating VLMs on counting with constraints in 3D generated images. It provides a compromise between highly artificial 2D images as used in VLMCountBench and real-world photos like TallyQA, which do not control for occlusions. We've presented strong indications that our dataset in the mixed version enables the finetuned VLM to better generalize to out-of-distribution images (TallyQA) by comparing results on our dataset with three other counting benchmarks and comparing the finetune on SITUATE with another one on the Pixmo count dataset of the same size.

We plan to increase the number of different object classes to include other objects with more relation to real life, but similar characteristics to the classes we already have. We think about cups/mugs and cans that would be a natural extension of cylinders, but could differ not only in the color, but also have photorealistic textures, e.g., for beverages. Similarly, spheres could be expanded to footballs, basketballs or other easily identifiable balls with rich texture. Furthermore, other objects that naturally come with many pieces to count like candles on a birthday cake or pieces of a chess game. This could narrow the gap between synthetic images on the one hand and realistic photos on the other hand. We should also vary the environment even more, e.g., including chairs or sofas instead of tables as reference points for the location-based questions. These measures would also lead to an increased number of image/question pairs that could be used to train larger models.

This work was carried out as part of the EU co-funded research project multi-modal man-machine interaction using AI (M4-SKI).

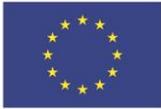